\documentclass[letterpaper, 10pt, conference]{ieeeconf} 

\IEEEoverridecommandlockouts 

\usepackage{cite}
\usepackage{amsmath,amssymb,amsfonts}
\usepackage{algorithm}
\usepackage[noend]{algorithmic} 
\usepackage{graphicx}
\usepackage{textcomp}
\usepackage{xcolor}
\def\BibTeX{{\rm B\kern-.05em{\sc i\kern-.025em b}\kern-.08em
    T\kern-.1667em\lower.7ex\hbox{E}\kern-.125emX}}
\usepackage{booktabs}
\usepackage{multirow}
\usepackage[super]{nth}
\usepackage{tikz}
\newcommand*\circled[1]{\tikz[baseline=(char.base)]{
            \node[shape=circle,draw,inner sep=0.2pt] (char) {#1};}}
\newcommand*\circledB[1]{\tikz[baseline=(char.base)]{
            \node[shape=circle,fill,inner sep=0.2pt] (char) {\textcolor{white}{#1}};}}
\usetikzlibrary{tikzmark}
\tikzset{circledColor/.style={circle,draw,inner sep=0.1em,line width=0.04em}}
\usepackage{soul}
\usepackage{setspace}
\usepackage{dblfloatfix}
\usepackage[urlcolor=blue]{hyperref}
\usepackage{cleveref}

\usepackage{fancyhdr}
\fancypagestyle{firstpage}{
  \fancyhf{}
  \chead{\small To appear at the 18th International Conference on Control, Automation, Robotics and Vision (ICARCV), December 2024, Dubai, UAE.}
  \cfoot{\thepage}
}

\title{\LARGE \bf
FastSpiker: Enabling Fast Training for Spiking Neural Networks on Event-based Data through Learning Rate Enhancements for Autonomous Embedded Systems
\vspace{-0.2cm}
}

\author{Iqra Bano*, Rachmad Vidya Wicaksana Putra*, Alberto Marchisio, Muhammad Shafique 
\thanks{Iqra Bano, Rachmad Vidya Wicaksana Putra, and Alberto Marchisio are with eBrain Lab, Division of Engineering, New York University (NYU) Abu Dhabi, United Arab Emirates;
{e-mail: \{ib2419, rachmad.putra, alberto.marchisio\}@nyu.edu}
}%
\thanks{Muhammad Shafique is the Director of eBrain Lab, Division of Engineering, New York University (NYU) Abu Dhabi, United Arab Emirates;
{e-mail: muhammad.shafique@nyu.edu}
}
\thanks{*These authors have equal contributions.}
}

\begin{document}

\maketitle
\pagestyle{plain}
\thispagestyle{firstpage}

\begin{spacing}{1}
    
\begin{abstract}
Autonomous embedded systems (e.g., robots) typically necessitate intelligent computation with low power/energy processing for completing their tasks. 
Such requirements can be fulfilled by embodied neuromorphic intelligence with spiking neural networks (SNNs) because of their high learning quality (e.g., accuracy) and sparse computation.
Here, the employment of event-based data is preferred to ensure seamless connectivity between input and processing parts. 
However, state-of-the-art SNNs still face a long training time to achieve high accuracy, thereby incurring high energy consumption and producing a high rate of carbon emission. 
Toward this, we propose \textit{FastSpiker}, a novel methodology that enables fast SNN training on event-based data through learning rate enhancements targeting autonomous embedded systems.
In FastSpiker, we first investigate the impact of different learning rate policies and their values, then select the ones that quickly offer high accuracy.
Afterward, we explore different settings for the selected learning rate policies to find the appropriate policies through a statistical-based decision. 
Experimental results show that our FastSpiker offers up to 10.5x faster training time and up to 88.39\% lower carbon emission to achieve higher or comparable accuracy to the state-of-the-art on the event-based automotive dataset (i.e., NCARS). 
In this manner, our FastSpiker methodology paves the way for green and sustainable computing in realizing embodied neuromorphic intelligence for autonomous embedded systems. 
\end{abstract}

\section{Introduction}
\label{Sec_Intro}

Autonomous embedded systems with tightly-constrained resources (e.g., robots, UAVs, and UGVs) typically require intelligent yet low power/energy processing for completing their data analytic tasks~\cite{Ref_Bartolozzi_EmbodiedNeuroIntel_Nature22}\cite{Ref_Putra_NeuromorphicAI_arXiv24}, such as object recognition, navigation, obstacle avoidance, and user assistance. 
These requirements can be satisfied by embodied neuromorphic artificial intelligence (AI) with spiking neural networks (SNNs), because SNNs offer high learning quality (e.g., accuracy) and sparse event-based computation~\cite{Ref_Roy_SpikeMachineIntel_Nature19, Ref_Bartolozzi_EmbodiedNeuroIntel_Nature22, Ref_Putra_NeuromorphicAI_arXiv24}. 
Therefore, the employment of event-based data (e.g., gesture dataset~\cite{Ref_Amir_DVSGesture_CVPR17} and cars dataset~\cite{Ref_Sironi_HATS_CVPR18}) is favorable to ensure seamless interfacing from input to processing parts~\cite{Ref_Putra_NeuromorphicAI_arXiv24}\cite{Ref_Bano_StudySNNParameters_arXiv24}. 

\begin{figure}[t]
\centering
\includegraphics[width=0.85\linewidth]{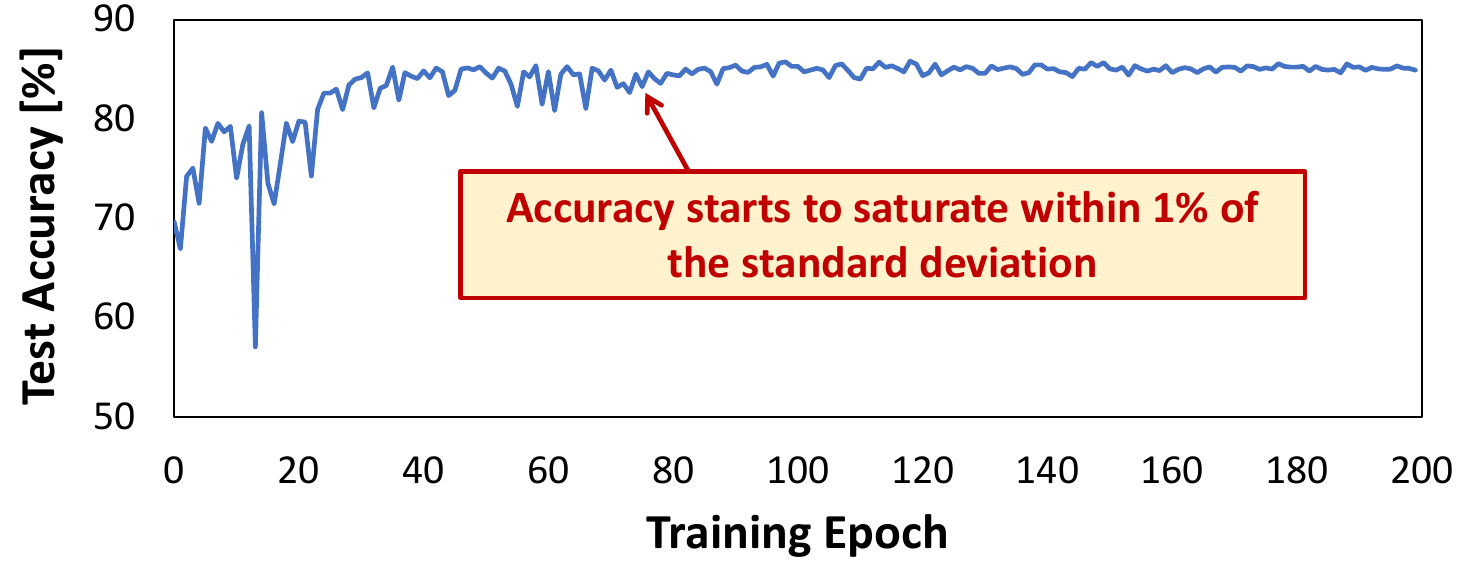}
\vspace{-0.3cm}
\caption{Accuracy profile of the state-of-the-art SNN model considering the NCARS dataset; adapted from~\cite{Ref_Viale_CarSNN_IJCNN21}. Its accuracy saturates within 1\% of the standard deviation after reaching at least around 80 training epochs.} 
\label{Fig_BaseAccuracy}
\vspace{-0.6cm}
\end{figure}

Currently, to achieve high accuracy in solving the given task, the existing SNN models still require a significantly long training time to adjust and fine-tune their parameter values. 
For instance, the state-of-the-art SNNs for processing event-based automotive data (i.e., NCARS dataset)~\cite{Ref_Viale_CarSNN_IJCNN21}\cite{Ref_Putra_SNN4Agents_FROBT24} require approximately 9 hours to complete the training phase with 200 epochs using the Nvidia RTX 6000 Ada GPU machines and achieve $\sim$85\% accuracy, as shown in Fig.~\ref{Fig_BaseAccuracy}. 
\textit{Such a long training time incurs huge energy consumption and emits a large amount of carbon}, which can negatively impact our environments~\cite{Ref_Strubell_CarbonEmission_ACL19, Ref_Strubell_EnergyConsideration_AAAI20, Ref_Putra_EnforceSNN_FNINS22}.
Moreover, a long training time also hinders the possibility of performing an efficient online training for updating the systems' knowledge at run-time in some application use-cases, such as systems that consider a continual learning scenario~\cite{Ref_Putra_NeuromorphicAI_arXiv24}\cite{Ref_Putra_SpikeDyn_DAC21}\cite{Ref_Putra_lpSpikeCon_IJCNN22}. 

The above discussion highlights the need for an efficient solution that enables fast SNN training while maintaining its high learning quality (i.e., accuracy).
Therefore, in this paper, \textit{\textbf{the targeted research problem} is how we can reduce the SNN training time while maintaining high accuracy}. 
An efficient solution to this problem may enable a fast and effective SNN training, and thereby reducing carbon emission for green and sustainable environments.

\subsection{State-of-the-art and Their Limitations}
\label{Sec_Intro_SOTA}

To minimize the training time of SNNs, the state-of-the-art work employed a direct timestep reduction with scaled parameter values during the training phase, thus curtailing the processing time of neurons while compensating the reduced spiking activity for maintaining the learning quality~\cite{Ref_Putra_TopSpark_IROS23}.
However, this work has not considered event-based data as input, thus limiting its energy efficiency gains and straightforward connectivity to event-based sensors, such as dynamic vision sensor (DVS) cameras. 
Besides such techniques, the reduction of SNN training time has not been explored further. 
Therefore, \textit{there is a significant need for investigating an alternate solution that can facilitate event-based input data for direct interfacing with event-based sensors}.

In conventional deep neural networks (DNNs), different learning rate (LR) policies have been explored to reduce their training time, such as exponential decay, one-cycle policy~\cite{Ref_Smith_NNhyperparameters_arXiv18}, cyclical policy~\cite{Ref_Smith_CyclicalLR_WACV17}, and warm restarts~\cite{Ref_Loshchilov_SGDR_arxiv16}\cite{Ref_Mishra_WarmRestart_TENCON19}. 
However, \textit{the impact of different LR policies in SNNs has not been explored, especially when considering event-based input data}. 
To illustrate the potential benefits of different LR policies, we perform an experimental case study, which is discussed in Section~\ref{Sec_Intro_CaseStudy}. 

\subsection{Motivational Case Study}
\label{Sec_Intro_CaseStudy}

\begin{figure}[t]
\centering
\includegraphics[width=0.95\linewidth]{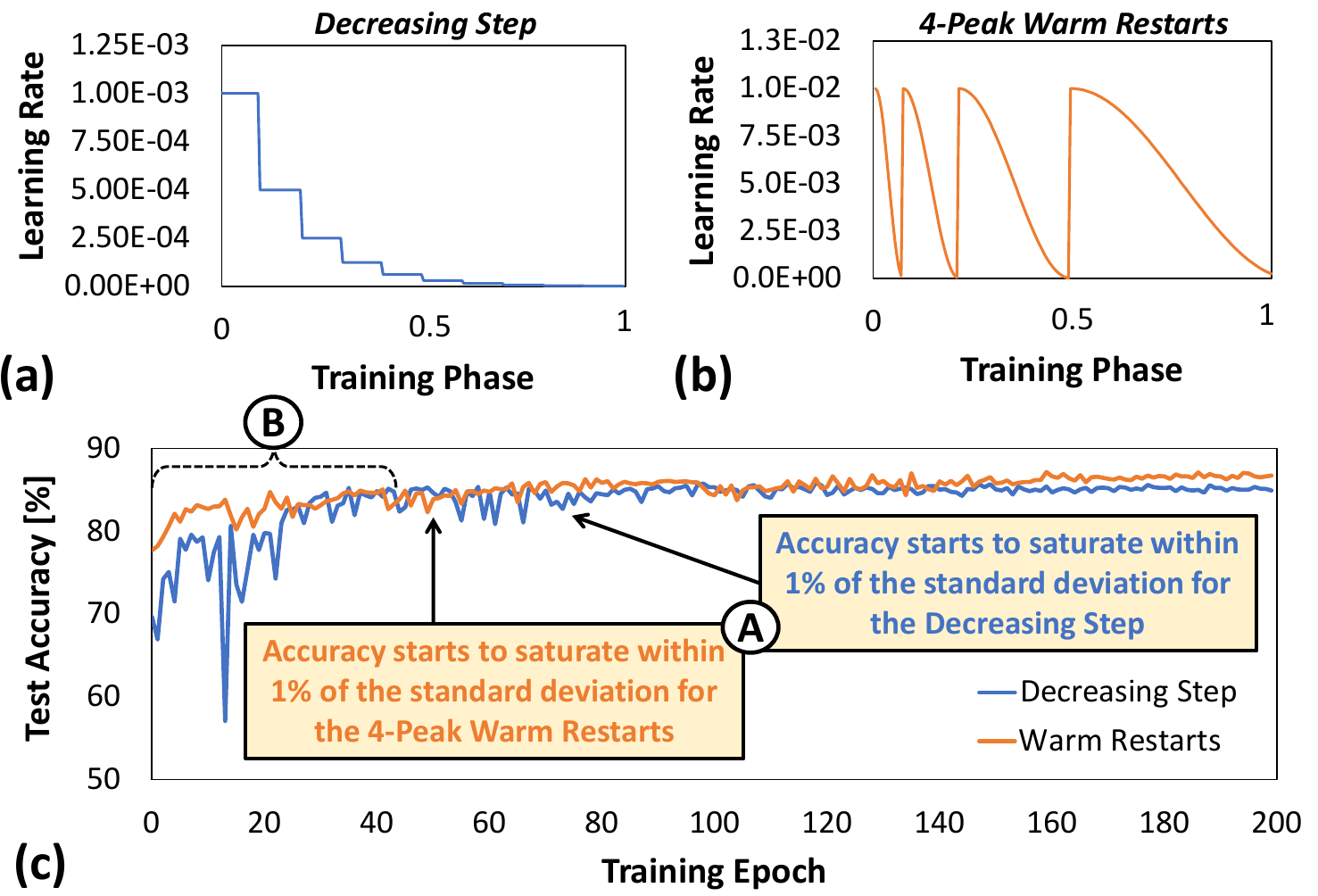}
\vspace{-0.3cm}
\caption{Learning rate policies considered in the case study: (a) decreasing step, and (b) 4-peak warm restarts. Note, if a training phase is defined as 200 training epochs, then the 0.5 training phase means 100 training epochs.  (c) Accuracy profiles from the decreasing step and warm restarts policies.} 
\label{Fig_CaseStudy}
\vspace{-0.5cm}
\end{figure}

We conduct a case study to investigate the impact of different LR policies on the accuracy. 
To do this, we consider two different LR policies, i.e., \textit{decreasing step} and \textit{warm restarts}, as shown in Fig.~\ref{Fig_CaseStudy}(a) and Fig.~\ref{Fig_CaseStudy}(b), respectively. 
For the network model, we consider a state-of-the-art SNN for the NCARS dataset from~\cite{Ref_Viale_CarSNN_IJCNN21} with 200 training epochs. 
Note, the details of LR policies are discussed in Section~\ref{Sec_Back_LearnPolicy}, while the experimental setup details are provided in Section~\ref{Sec_EvalMethod}.
The experimental results are shown in Fig.~\ref{Fig_CaseStudy}(c), from which we make several key observations, as follows.
\begin{itemize}
    \item Different LR policies may need different numbers of training epochs to reach stability in their learning curves. 
    For instance, the decreasing step policy achieves stable accuracy scores (i.e., having less than 1\% of standard deviation) after 77 training epochs, while the warm restarts policy achieves this faster with 50 training epochs; see \circled{A}.   
    \item Different initial LR values may lead to different responses in terms of learning curves at the early training phase. 
    For instance, the decreasing step policy faces significant fluctuation\footnote{Fluctuation is defined as the changes of accuracy across multiple epochs in the training phase. Fluctuation is considered significant if the accuracy change exceeds a stability threshold value ($acc_{th}$). In this work, $acc_{th}$ is set as 1\% of standard deviation of the accuracy scores from the last 10 training epochs, following the criteria in~\cite{Ref_Putra_EnforceSNN_FNINS22}.} at the early training phase, while the warm restarts policy faces less fluctuation; see \circled{B}.
    This insight is essential for devising a policy that can train the SNNs faster. 
\end{itemize}

These observations indicate that there is an opportunity to improve the training time of SNNs on event-based data by employing an effective LR policy. 
However, developing such a policy is a non-trivial task, as it imposes the following research challenges. 
\begin{itemize}
    \item The LR policy should provide effective learning even at the early training phase to enable fast training, thereby having minimum fluctuation in the learning curve.
    \item To find an appropriate LR policy, a systematic methodology is needed to explore different LR policies and quantitatively select the ones that offer high accuracy with fast training time. 
\end{itemize} 

\subsection{Our Novel Contributions}
\label{Sec_Intro_Novelty}

\begin{figure}[hbtp]
\vspace{-0.3cm}
\centering
\includegraphics[width=0.95\linewidth]{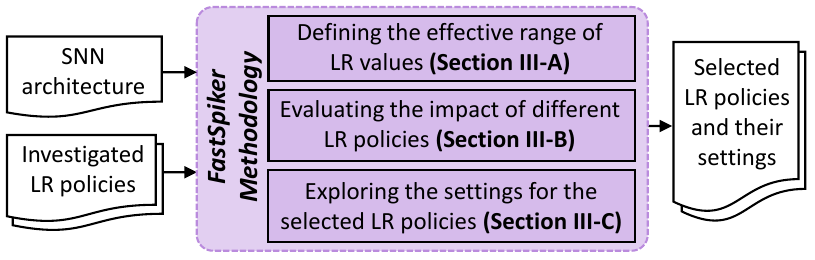}
\vspace{-0.3cm}
\caption{Our novel contributions in this paper, highlighted in purple.} 
\label{Fig_Novelty}
\vspace{-0.6cm}
\end{figure}

To solve the above research challenges, we propose \textit{\textbf{FastSpiker}, a novel methodology that enables \underline{Fast} training for \underline{Spik}ing neural networks on \underline{e}vent-based input data through learning \underline{r}ate enhancements}; see an overview in Fig.~\ref{Fig_Novelty}.
The key steps of our FastSpiker methodology are the following.
\begin{itemize}
    \item \textbf{Determining the effective range of LR values (Section \ref{Sec_FastSpiker_Range}):} 
    It finds the effective range of LR to learn the given event-based data through experiments that explore the impact of different LR values using the baseline policy on the accuracy.
    \item \textbf{Evaluating the impact of different LR policies (Section \ref{Sec_FastSpiker_Evaluate}} 
    It evaluates the impact of different widely-known LR policies (e.g., \textit{decreasing step}, \textit{exponential decay}, \textit{one-cycle}, \textit{cyclical}, \textit{decreasing cyclical}, and \textit{warm restarts}) for learning event-based data, then select the ones that offer high accuracy.
    \item \textbf{Exploring the settings for the selected LR policies (Section \ref{Sec_FastSpiker_Explore}):} 
    It explores different possible settings for the selected LR policies to find the appropriate settings through a statistical variance-based decision. 
\end{itemize}

\textbf{Key Results:}
To evaluate our FastSpiker methodology, we employ Python-based implementations considering NCARS dataset, then run them on the Nvidia RTX 6000 Ada machines.
Experimental results show that our FastSpiker can expedite training time by up to 10.5x and reduce carbon emission by up to 88.39\% for achieving higher or comparable accuracy scores than the state-of-the-art. 

\begin{figure*}[t]
\centering
\includegraphics[width=\linewidth]{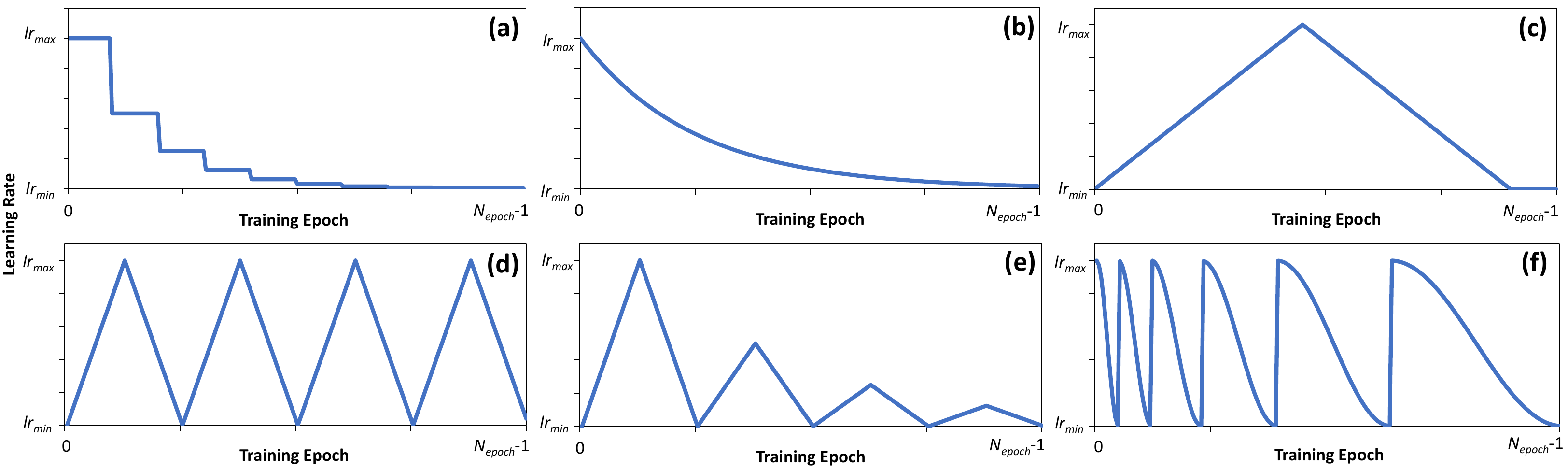}
\vspace{-0.7cm}
\caption{Different types of LR policies: (a) decreasing step, (b) exponential decay, (c) one-cycle, (d) cyclical, (e) decreasing cyclical, and (f) warm restarts.} 
\label{Fig_LRPolicies}
\vspace{-0.5cm}
\end{figure*}
\smallskip

\section{Background}
\label{Sec_Back}

\subsection{Spiking Neural Networks (SNNs)}
\label{Sec_Back_SNNs}

Modeled after the neural process of the human brain, spiking neural networks (SNNs) adopt how neurons transmit spikes for data communication and computation. 
This enables SNNs to excel in scenarios where power/energy efficiency and real-time processing are paramount, such as autonomous embedded systems (e.g., robotics, UAVs, and UGVs). 
SNNs receive spike sequences as input, and process the information using spiking neuron models. 
One of the most widely-used neuron model in the SNN community is the Leaky Integrate-and-Fire (LIF), whose neuronal behavior can be stated as follows.
\begin{equation}
    \begin{split}
    \frac{dV_{m}}{dt} & = \frac{1}{\tau}(-(V_{m}-V_{r})+RI) \\
    \text{if} \;\;  & V_{m} \geq V_{th} \;\; \text{then} \;\; V_{m} \leftarrow V_{r} \\
    \end{split}
    \label{Eq_LIF}
\end{equation}
$V_{m}$, $V_{r}$, and $V_{th}$ denote the neurons' membrane, reset, and threshold potentials, respectively. 
$R$ and $I$ denote the input resistance and current (spike), respectively. 
Meanwhile, $\tau$ denotes to the time constant of $V_{m}$ decay.  
In this work, the SNN architecture is based on studies in~\cite{Ref_Viale_CarSNN_IJCNN21}\cite{Ref_Putra_SNN4Agents_FROBT24} with an input sample with 100x100 pixels. 

Training SNNs in a supervised setting faces non-differentiable issues when computing the loss function~\cite{Ruckauer_2019arxiv_NonDifferentiableLossFunctionSNN}. 
To overcome this, two possible supervised-based training solutions have been proposed in the literature: \textit{DNN-to-SNN conversion} and \textit{direct SNN training with surrogate gradient}~\cite{Ref_Putra_NeuromorphicAI_arXiv24}. 
Previous works showed that the DNN-to-SNN conversion offers sub-optimal performance compared to training directly in the spiking domain via a surrogate gradient-based learning rule~\cite{Ref_Kim_SNASNet_ECCV22, Ref_Putra_SpikeNAS_arxiv24}.
In this category, Spatio-Temporal Back-Propagation (STBP)~\cite{Ref_Wu_STBP_FNINS18} is a prominent technique, hence in this work, we employ the STBP for the selected SNN architecture. 

\subsection{Learning Rate Policies}
\label{Sec_Back_LearnPolicy}

Various LR policies have been proposed in the conventional DNN domain, such as the decreasing step, exponential decay, one-cycle, cyclical, decreasing cyclical, and warm restarts policies~\cite{Ref_Smith_NNhyperparameters_arXiv18, Ref_Smith_CyclicalLR_WACV17, Ref_Loshchilov_SGDR_arxiv16, Ref_Mishra_WarmRestart_TENCON19}. 
These investigated LR policies are illustrated in Fig.~\ref{Fig_LRPolicies}, and their descriptions are provided in the following. 

\subsubsection{Decreasing Step}

It optimizes SNNs by gradually reducing LR values during the training phase in the step fashion, as shown in Fig.~\ref{Fig_LRPolicies}(a). 
In this work, LR decreases by a factor of 0.5 every 20 epochs following~\cite{Ref_Viale_CarSNN_IJCNN21}\cite{Ref_Putra_SNN4Agents_FROBT24}. 
This systematic reduction is to fine-tune the parameters, improving convergence over time.

\subsubsection{Exponential Decay}

It optimizes SNNs by gradually reducing LR values during the training phase every epoch, as shown in Fig.~\ref{Fig_LRPolicies}(b). 
In this work, LR decreases by a factor of 0.98 every epoch. 
This systematic reduction is also expected to fine-tune the parameters, improving convergence over time smoothly.

\subsubsection{One-Cycle}

It optimizes SNNs by realizing a single cycle of LR values, where the LR increases to the maximum value, and then decreases back to the minimum one~\cite{Ref_Smith_NNhyperparameters_arXiv18}; see Fig.~\ref{Fig_LRPolicies}(c). 
In this work, the initial LR is set as 1e-5, then it increases to 1e-2 by the 100th epoch. 
From the 100th to the 180th epoch, it linearly reduces the rate back to 1e-5 for fine-tuning. 
Then, the last 20 epochs are dedicated for stabilizing the accuracy at a low LR. 

\subsubsection{Cyclical}

It optimizes SNNs by realizing cyclical LR values that oscillate within upper and lower bounds, aiding in escaping local minima for better solutions~\cite{Ref_Smith_CyclicalLR_WACV17}; see Fig.~\ref{Fig_LRPolicies}(d). 
Practically, with a minimum LR of 1e-5 and a maximum LR of 1e-2, and the number of cycles can be set to 4. 
Within each cycle, the LR linearly increases from the minimum value to the maximum one during the first half (25 epochs), then decreases back to the minimum value during the second half (another 25 epochs).

\subsubsection{Decreasing Cyclical}

It optimizes SNNs by realizing cyclical LR values that oscillate between pre-defined maximum and minimum values over several cycles, with the maximum LR gradually decreases with each cycle; see Fig.~\ref{Fig_LRPolicies}(e).
Practically, the maximum LR for the next cycle is 0.9x of the current maximum LR value. This progressive reduction helps in fine-tuning the model. 

\subsubsection{Warm Restarts}

It optimizes SNNs by periodically resetting LR to a high value after it decreases over a number of epochs, enforcing a fresh restart that helps escaping a local minima to find a global minima~\cite{Ref_Loshchilov_SGDR_arxiv16}\cite{Ref_Mishra_WarmRestart_TENCON19}; see Fig.~\ref{Fig_LRPolicies}(f).
Practically, with a minimum LR of 1e-5 and a maximum LR of 1e-2, the training phase can be divided into several cycles with different lengths. 

\subsection{Event-based Data}
\label{Sec_Back_EventData}

\begin{figure}[h]
\vspace{-0.3cm}
\centering
\includegraphics[width=\linewidth]{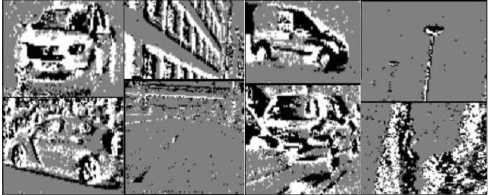}
\vspace{-0.6cm}
\caption{Illustration of the NCARS dataset~\cite{Ref_Sironi_HATS_CVPR18}~\cite{Ref_Viale_CarSNN_IJCNN21}.} 
\label{Fig_NCARS_dataset}
\vspace{-0.4cm}
\end{figure}

In this work, we consider the Prophesees' NCARS dataset~\cite{Ref_Sironi_HATS_CVPR18}. 
This dataset consists of about 24K samples, each having a 100ms duration. 
It contains either ``car'' or ``background'' samples, which are captured through recordings using the event-based camera~\cite{Ref_Sironi_HATS_CVPR18}; see Fig.~\ref{Fig_NCARS_dataset}.
Each input sample has a sequence of events that contain information of the spatial coordinates of the pixel, the timestamp of the event, and the polarity of the brightness which is either positive or negative. 
In this work, we employ $100 \times 100$ pixels since it is sufficient to provide important information yet small in size~\cite{Ref_Putra_SNN4Agents_FROBT24}.

\vspace{-0.1cm}
\subsection{Carbon Emission of NN Training}

Recent works studied that training neural networks (NNs) emits a huge amount of carbon, hence raising environmental concerns that increases the rates of natural disasters~\cite{Ref_Strubell_CarbonEmission_ACL19}\cite{Ref_Strubell_EnergyConsideration_AAAI20}.
Hence, reducing the NN training time can minimize the carbon emission, preserving green and sustainable environments~\cite{Ref_Putra_EnforceSNN_FNINS22}.
Previous work~\cite{Ref_Strubell_CarbonEmission_ACL19} proposed Eq.~\ref{Eq_Carbon1}-\ref{Eq_Carbon2} to estimate the carbon emission from NN training. 
\begin{equation}
    CO_2e = 0.954 \cdot P_{train} \cdot t
    \label{Eq_Carbon1}
\end{equation}
\vspace{-0.2cm}
\begin{equation}
    P_{train} = \frac{1.58 \cdot (P_{cpu}+P_{mem}+g \cdot P_{gpu})}{1000}
    \label{Eq_Carbon2}
    \smallskip
\end{equation}

Here, $CO_2e$ is the estimated carbon emission (i.e., $CO_2$), which is a function of the total training power ($P_{train}$) and the training duration ($t$).
Meanwhile, $P_{cpu}$ represents the average power from CPUs, $P_{mem}$ represents the average power from main memories (DRAMs), $P_{gpu}$ represents the average power from a GPU, and $g$ represents the number of GPUs for training.

\section{FastSpiker Methodology}
\label{Sec_FastSpiker}

\begin{figure}[h]
\vspace{-0.2cm}
\centering
\includegraphics[width=\linewidth]{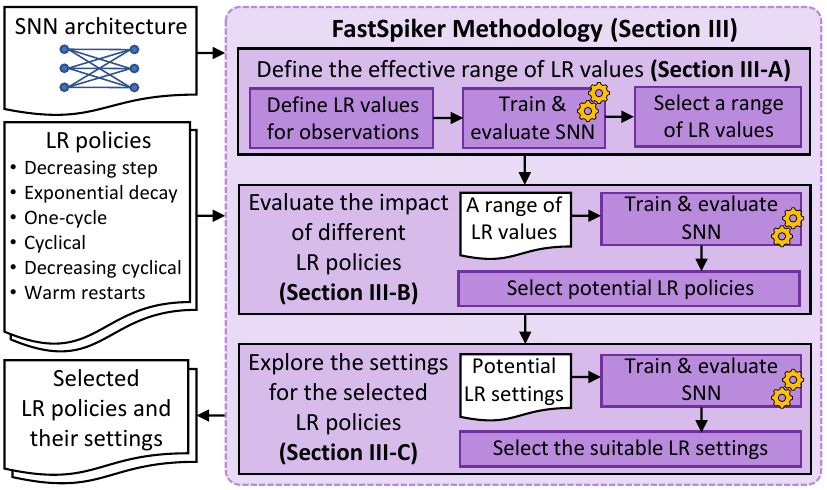}
\vspace{-0.6cm}
\caption{The FastSpiker methodology, showing the novel key steps in purple.} 
\label{Fig_FastSpiker}
\vspace{-0.4cm}
\end{figure}

FastSpiker methodology aims to systematically find the effective LR policies that can lead the given SNN to quickly learn from event-based input data, thereby achieving high accuracy within a relatively short training time.  
Our FastSpiker employs several steps as shown in Fig.~\ref{Fig_FastSpiker}, which are discussed in Sections~\ref{Sec_FastSpiker_Range}-\ref{Sec_FastSpiker_Explore}. 

\vspace{-0.1cm}
\subsection{Determination of the Effective Range of LR Values}
\label{Sec_FastSpiker_Range}

\begin{figure}[h]
\vspace{-0.2cm}
\centering
\includegraphics[width=\linewidth]{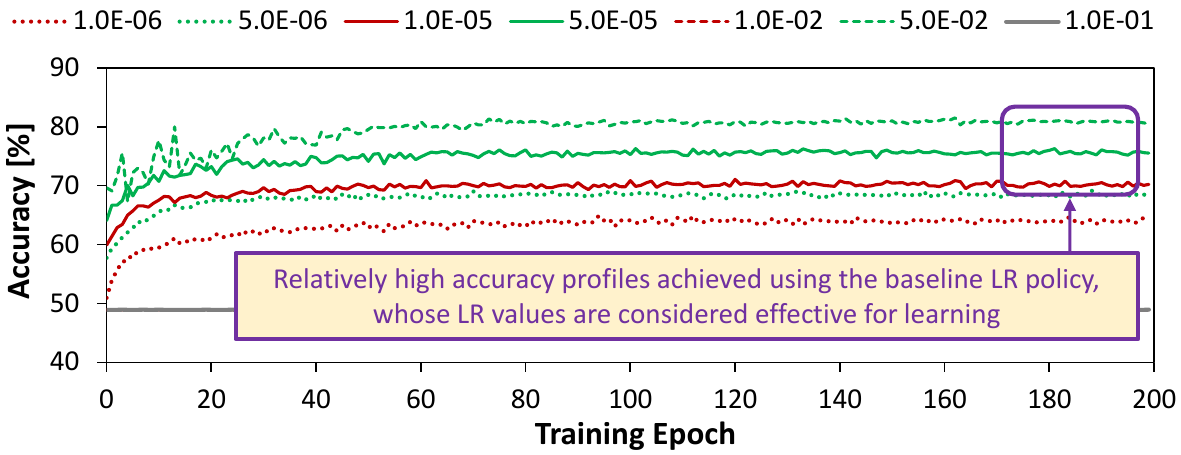}
\vspace{-0.6cm}
\caption{Accuracy profiles obtained using the state-of-the-art LR policy for learning event-based data (i.e., decreasing step) considering different LR values: $1\mathrm{e}{\text{-}6}$, $5\mathrm{e}{\text{-}6}$, $1\mathrm{e}{\text{-}5}$, $5\mathrm{e}{\text{-}5}$, $1\mathrm{e}{\text{-}2}$, $5\mathrm{e}{\text{-}2}$, and $1\mathrm{e}{\text{-}1}$.} 
\label{Fig_RangeSweep}
\vspace{-0.2cm}
\end{figure}

Each LR policy requires a pre-defined range of values to tailor its behavior, in the manner that the respective LR policy makes the network effectively learn the event-based input data.
To determine the appropriate range of values, a simple yet effective way is by performing observations through network training that considers different LR values from small to large ones (i.e., $1\mathrm{e}{\text{-}6}$, $5\mathrm{e}{\text{-}6}$, $1\mathrm{e}{\text{-}5}$, $5\mathrm{e}{\text{-}5}$, $1\mathrm{e}{\text{-}2}$, $5\mathrm{e}{\text{-}2}$, and $1\mathrm{e}{\text{-}1}$), and the results are shown in Fig.~\ref{Fig_RangeSweep}.

These results show that the LR value should not be too small (e.g., $1\mathrm{e}{\text{-}6}$) or too big (i.e., $1\mathrm{e}{\text{-}1}$), since they lead to ineffective learning process.
Therefore, \textit{in this work, we consider the range of LR values between $1\mathrm{e}{\text{-}5}$ and $1\mathrm{e}{\text{-}2}$}.

\subsection{Evaluation of the Impact of Different LR Policies}
\label{Sec_FastSpiker_Evaluate}

This step aims at evaluating the impact of different LR policies for learning event-based data, while employing the selected range of values.  
To do this, we first perform network training for each investigated LR policy on the NCARS dataset.
Afterward, policies that offer high accuracy will be selected as the potential solutions. 
To select the appropriate solutions, we employ \textit{a statistical-based decision to determine if the training phase is considered complete and can be terminated}.
Here, we consider the training phase complete when the accuracy scores do not exceed the defined stability threshold value ($acc_{th}$) anymore, thereby indicating that the learning curves have saturated and reached stability. 
Specifically, we set the $acc_{th}$ as 1\% standard deviation of the accuracy scores from the last 10 training epochs, following the criteria in~\cite{Ref_Putra_EnforceSNN_FNINS22}.

In this work, we evaluate the impact of well-known LR policies, including the \textit{decreasing step}, \textit{exponential decay}, \textit{one-cycle}, \textit{cyclical}, \textit{decreasing cyclical}, and \textit{warm restarts}.  
Here, we employ the following parameter settings for all investigated LR policies:
\begin{itemize}
    \item Initial learning rate ($lr$) = $1\mathrm{e}{\text{-}3}$
    \item Batch size = 40 samples
    \item Neurons' membrane threshold potential ($V_{th}$) = 0.4
    \item Details of the experimental setup are provided in Section~\ref{Sec_EvalMethod}
\end{itemize}
Meanwhile, other settings that are applicable to specific LR policies follow the pseudo-codes in Alg.~\ref{Alg_DecreasingStep}-Alg.\ref{Alg_WarmRestarts}. 

\begin{algorithm}[t!]
\caption{Decreasing Step LR Policy}
\label{Alg_DecreasingStep}
\footnotesize
\textbf{Require:} optimizer, initial LR ($init=0.1$), reduction factor ($r\_f=0.98$), reduction interval ($r\_int=20$), \#epochs ($epochs=200$);
\begin{spacing}{1}
\begin{algorithmic}[1]
  \STATE $cur\_lr \gets init$ // initialization
  \FOR{$ep$ = 1 to $epochs$}
    \IF{$ep$ \% $r\_int$ == 0}
      \STATE $cur\_lr \gets cur\_lr \times r\_f$
    \ENDIF
    \STATE Update $cur\_lr$ in optimizer
  \ENDFOR
  \RETURN Updated optimizer
\end{algorithmic}
\end{spacing}
\end{algorithm}
\begin{algorithm}[t!]
\caption{Exponential Decay LR Policy}
\label{Alg_ExpDecay}
\footnotesize
\textbf{Require:} optimizer, initial LR ($init=0.1$), \#epochs ($epochs=200$), decay rate ($d\_rate=0.98$), decay steps ($d\_steps=1$);
\begin{spacing}{1}
\begin{algorithmic}[1]
\STATE $lr \gets init$ // initialization
\FOR{$ep$ = 1 to $epochs$}
  \STATE $lr \gets init \times d\_rate^{(\frac{ep}{d\_steps})}$
  \STATE Update $lr$ in optimizer
\ENDFOR
\RETURN Updated optimizer
\end{algorithmic}
\end{spacing}
\end{algorithm} 
\setlength{\textfloatsep}{6pt}
\begin{algorithm}[t!]
\caption{One-Cycle LR Policy}
\label{Alg_OneCycle}
\footnotesize
\textbf{Require:} optimizer, initial LR ($init=0.1$), \#epochs ($epochs=200$), peak epoch ($p\_ep=90$), drop epoch ($d\_ep=180$), start LR ($start=1\mathrm{e}{\text{-}5}$), max LR ($max=1\mathrm{e}{\text{-}2}$), min LR ($min=1\mathrm{e}{\text{-}5}$), end LR ($end=1\mathrm{e}{\text{-}8}$);
\begin{spacing}{1}
\begin{algorithmic}[1]
\STATE $lr \gets init$ // initialization
\FOR{$ep$ = 1 to $epochs$}
  \IF{$ep$ $<$ $p\_ep$}
    \STATE $lr \gets start + \frac{(max - start)}{p\_ep} \times ep$
  \ELSIF{$ep$ $<$ $d\_ep$}
    \STATE $lr \gets max - \frac{(max - min)}{(d\_ep - p\_ep)} \times (ep - p\_ep)$
  \ELSE
    \STATE $lr \gets min - \frac{(min - end)}{(epochs - d\_ep)} \times (ep - d\_ep)$
  \ENDIF
  \STATE Update $lr$ in optimizer 
\ENDFOR
\RETURN Updated optimizer
\end{algorithmic}
\end{spacing}
\end{algorithm}
\setlength{\textfloatsep}{2pt}
\begin{algorithm}[t!]
\caption{Cyclical LR Policy}
\label{Alg_Cyclical}
\footnotesize
\textbf{Require:} optimizer, initial LR ($init=0.1$), \#epochs ($epochs=200$), min LR ($min=1\mathrm{e}{\text{-}5}$), max LR ($max=1\mathrm{e}{\text{-}2}$), half cycle ($h\_cycle=25$);
\begin{spacing}{1}
\begin{algorithmic}[1]
  \STATE $lr \gets init$ // initialization
  \FOR{$ep$=1 to $epochs$}
    \STATE $cycle\_pos \gets ep \% (h\_cycle \times 2)$
    \IF{$cycle\_pos < h\_cycle$}
      \STATE $lr \gets min + \frac{(max - min) \times cycle\_pos}{h\_cycle}$
    \ELSE
      \STATE $lr \gets max - \frac{(max - min) \times (cycle\_pos - h\_cycle)}{h\_cycle}$
    \ENDIF
    \STATE Update $lr$ in optimizer
  \ENDFOR
\RETURN Updated optimizer
\end{algorithmic}
\end{spacing}
\end{algorithm}
\setlength{\textfloatsep}{2pt}
\begin{algorithm}[t!]
\caption{Decreasing Cyclical LR Policy}
\label{Alg_DecreasingCyclical}
\footnotesize
\textbf{Require:} optimizer, initial LR ($init=0.1$), number of epochs ($epochs=200$), max LR ($max=1\mathrm{e}{\text{-}2}$), min LR ($min=1\mathrm{e}{\text{-}5}$), cycle length ($c\_length=40$);
\begin{spacing}{1}
\begin{algorithmic}[1]
\STATE $lr \gets init$ // initialization
\FOR{$ep$= 1 to $epochs$}
  \STATE $N\_cycles \gets \left\lfloor \frac{ep}{c\_length} \right\rfloor$
  \STATE $lr\_dec \gets \frac{(max - min)}{\left(\frac{epochs}{c\_length}\right) - 1}$

  \STATE $cur\_max \gets max - (lr\_dec \times N\_cycles)$
  \STATE $cur\_prog \gets \frac{ep \% c\_length}{c\_length}$
  
  \STATE $lr \gets cur\_max - (cur\_max - min) \times cur\_prog$
  \STATE $lr \gets \max(lr, min)$
  \STATE Update $lr$ in optimizer
\ENDFOR
\RETURN Updated optimizer
\end{algorithmic}
\end{spacing}
\end{algorithm}
\setlength{\textfloatsep}{2pt}
\begin{algorithm}[t!]
\caption{Warm Restarts LR Policy}
\label{Alg_WarmRestarts}
\footnotesize
\textbf{Require:} optimizer, initial LR ($init=0.1$), number of epochs ($epochs=200$), maximum initial period ($T_{max}=4$), min LR ($min=1\mathrm{e}{\text{-}5}$), period multiplier ($T_{mult}=2$);
\begin{spacing}{1}
\begin{algorithmic}[1]
\STATE $lr \gets init$ // initialization
\FOR{$ep$ = 1 to $epochs$}
  \STATE $T_i \gets T_{max}$
  \STATE $t_{cur} \gets ep$
  \FOR{i = 0 to $ep$}
    \IF{$t_{cur} < T_i$}
      \STATE \textbf{break}
    \ENDIF
    \STATE $t_{cur} \gets t_{cur} - T_i$
    \STATE $T_i \gets T_i \times T_{mult}$
  \ENDFOR
  \STATE $lr \gets min + (lr - min) \times 0.5 \times (1 + \cos(\pi \times \frac{t_{cur}}{T_i}))$
  \STATE Update $lr$ in optimizer
\ENDFOR
\RETURN Updated optimizer
\end{algorithmic}
\end{spacing}
\end{algorithm} 
\setlength{\textfloatsep}{2pt}

\begin{figure*}[!b]
\vspace{-0.3cm}
\centering
\includegraphics[width=0.9\linewidth]{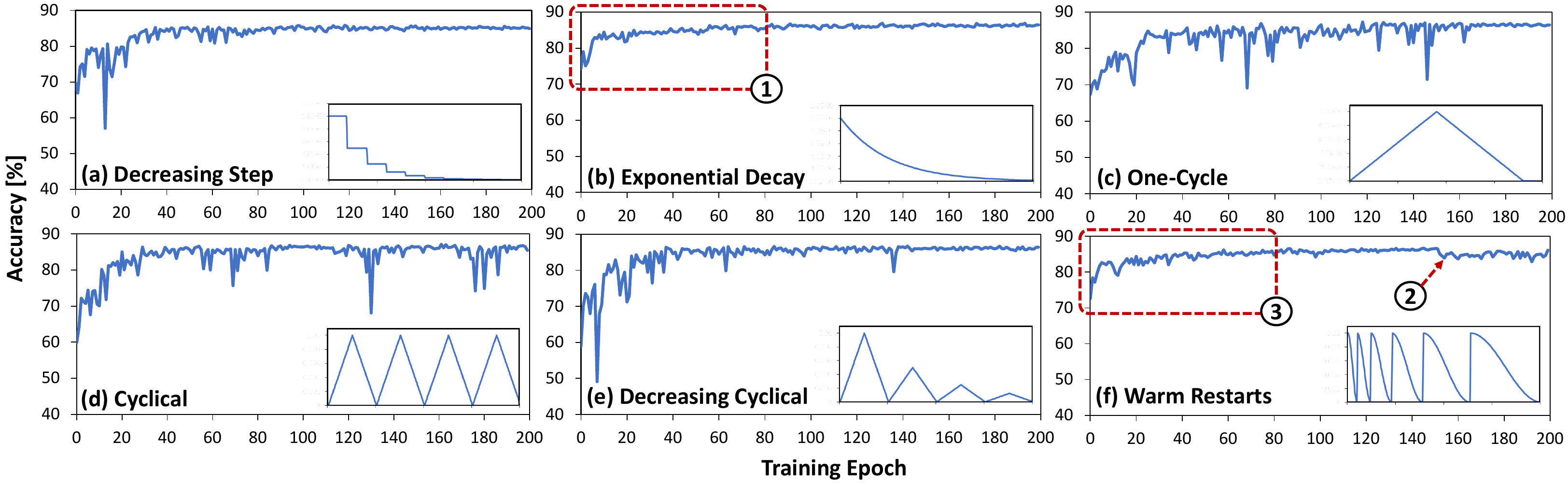}
\vspace{-0.3cm}
\caption{Accuracy profiles showing the learning curves of the SNN model when employing different LR policies: (a) decreasing step, (b) exponential decay, (c) one-cycle, (d) cyclical, (e) decreasing cyclical, and (f) warm restarts.} 
\label{Fig_AccuracyLRPolicies}
\vspace{-0.3cm}
\end{figure*}

The experimental results are shown in Fig.~\ref{Fig_AccuracyLRPolicies}, from which we make following observations.
\begin{itemize}
    \item The decreasing step policy leads to a stable learning curve after 77 training epochs. 
    It is faster as compared to the one-cycle, cyclical, and decreasing cyclical policies, which lead to stable learning curves after 173, 197, and 147 training epochs, respectively. 
    Therefore, \textit{the one-cycle, cyclical, and decreasing cyclical policies are not considered as solution candidates}. 
    \item The exponential decay policy leads to a stable learning curve after 34 training epochs, which is faster than other policies. 
    Moreover, it also has relatively small fluctuation in the early training phase as shown in \circled{1}. 
    Therefore, \textit{this policy is considered as a solution candidate}. 
    \item The warm restarts policy with 6 peaks leads to a stable learning curve after 156 training epochs; see \circled{2}.
    However, overall, it has relatively small fluctuation in the early training phase; see \circled{3}.
    Meanwhile, 4-peaks warm restarts policy leads to a stable learning curve after 51 training epochs. 
    This shows that the settings of warm restarts can be adjusted further to reach stable accuracy faster. 
    Therefore, \textit{this policy is also considered as a solution candidate}. 
\end{itemize}
Based on these observations,\textit{we choose the exponential decay and warm restarts policies as the selected the potential solutions}, whose settings will be explored and discussed further in Section~\ref{Sec_FastSpiker_Explore}. 

\subsection{Exploration of the Settings for Selected LR Policies}
\label{Sec_FastSpiker_Explore}

To quickly maximize the learning quality (i.e., accuracy), the selected LR policies (i.e., exponential decay and warm restarts) require to employ effective parameter settings, such as the \textit{initial learning rate ($init\_lr$)}, \textit{batch size ($B$)}, and \textit{neurons' threshold potential ($V_{th}$)}. 
However, finding the appropriate values is also a non-trivial task. 
Toward this, \textit{we explore different combinations of potential values for the investigated parameters}.
Its key idea is to vary these values (i.e., especially for $B$ and $V_{th}$ as summarized in Table~\ref{Tab_Explore}) and observe their impact on accuracy. 
Here, WR\_\textit{x}P refers to the warm restarts policy with \textit{x} peaks during a training phase, and ED\_B\textit{p}\_V\textit{q} refers to the warm restarts policy with \textit{p} batch size and \textit{0.q} threshold potential during a training phase.

\begin{table}[h]
\centering
\scriptsize
\caption{Combinations of values for the investigated parameters.} 
\vspace{-0.1cm}
\label{Tab_Explore}
\begin{tabular}{c|c|c|c|ll|c|c|c|}
\cline{2-4} \cline{7-9}
\textbf{} & \textbf{$init\_lr$} & \textbf{$B$} & \textbf{$V_{th}$} & & & \textbf{$init\_lr$} & \textbf{$B$} & \textbf{$V_{th}$} \\ \cline{1-4} \cline{6-9} 
\multicolumn{1}{|c|}{\textbf{\begin{tabular}[c]{@{}c@{}}WR\_2P\\ WR\_3P\\ WR\_4P\\ WR\_6P\\ WR\_7P\\ WR\_10P\end{tabular}}} & \begin{tabular}[c]{@{}c@{}}1e-2\\ 1e-2\\ 1e-2\\ 1e-2\\ 1e-2\\ 1e-2\end{tabular} & \begin{tabular}[c]{@{}c@{}}40\\ 40\\ 40\\ 40\\ 40\\ 40\end{tabular} & \begin{tabular}[c]{@{}c@{}}0.4\\ 0.4\\ 0.4\\ 0.4\\ 0.4\\ 0.4\end{tabular} & \multicolumn{1}{l|}{} & \multicolumn{1}{c|}{\textbf{\begin{tabular}[c]{@{}c@{}}ED\_B40\_V3\\ ED\_B40\_V4\\ ED\_B40\_V5\\ ED\_B40\_V6\\ ED\_B30\_V4\\ ED\_B20\_V4\end{tabular}}} & \begin{tabular}[c]{@{}c@{}}1e-2\\ 1e-2\\ 1e-2\\ 1e-2\\ 1e-2\\ 1e-2\end{tabular} & \begin{tabular}[c]{@{}c@{}}40\\ 40\\ 40\\ 40\\ 30\\ 20\end{tabular} & \begin{tabular}[c]{@{}c@{}}0.3\\ 0.4\\ 0.5\\ 0.6\\ 0.4\\ 0.4\end{tabular} \\ \cline{1-4} \cline{6-9} 
\end{tabular}
\vspace{0.1cm}
\end{table}

\section{Evaluation Methodology}
\label{Sec_EvalMethod}

\begin{figure}[h]
\vspace{-0.3cm}
\centering
\includegraphics[width=\linewidth]{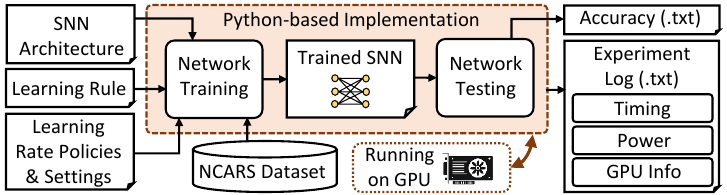}
\vspace{-0.7cm}
\caption{Experimental setup in the evaluation.} 
\label{Fig_EvalMethod}
\vspace{-0.3cm}
\end{figure}

Fig~\ref{Fig_EvalMethod} shows the experimental setup for evaluating our FastSpiker methodology. 
We employ a Python-based implementation that runs on Nvidia RTX 6000 Ada GPU machines, and the generated outputs are the accuracy and the log of experiments, including accuracy, processing time, and power consumption of the SNN training.
Additional experimental logs like GPU information (e.g., number of devices) are also recorded for estimating the carbon emission. 
For this calculation, we employ Eq.~\ref{Eq_Carbon1}-\ref{Eq_Carbon2} and focus on the carbon emission from GPU processes. 
For the network, we employ the SNN architecture with the STBP learning rule and 200 epochs for a training phase, following~\cite{Ref_Viale_CarSNN_IJCNN21}. 
Here, the NCARS dataset is used as the workload. 
Furthermore, we consider the state-of-the-art LR policy for event-based data (i.e., decreasing step with $init\_lr$=1e-3, $B$=40, and $V_{th}$=0.4) as the comparison partner. 
For brevity, this policy is referred to as ``SOTA\_DS''.
Furthermore, we set the $acc_{th}$ as 1\% standard deviation of the accuracy from the last 10 training epochs, for evaluating the stability of the learning curves.

\vspace{-0.2cm}
\section{Experimental Results}
\label{Sec_Results}

\subsection{Accuracy Profiles and the Reduction of Training Time}
\label{Sec_Results_TriningTime}

\begin{figure}[h]
\vspace{-0.4cm}
\centering
\includegraphics[width=0.95\linewidth]{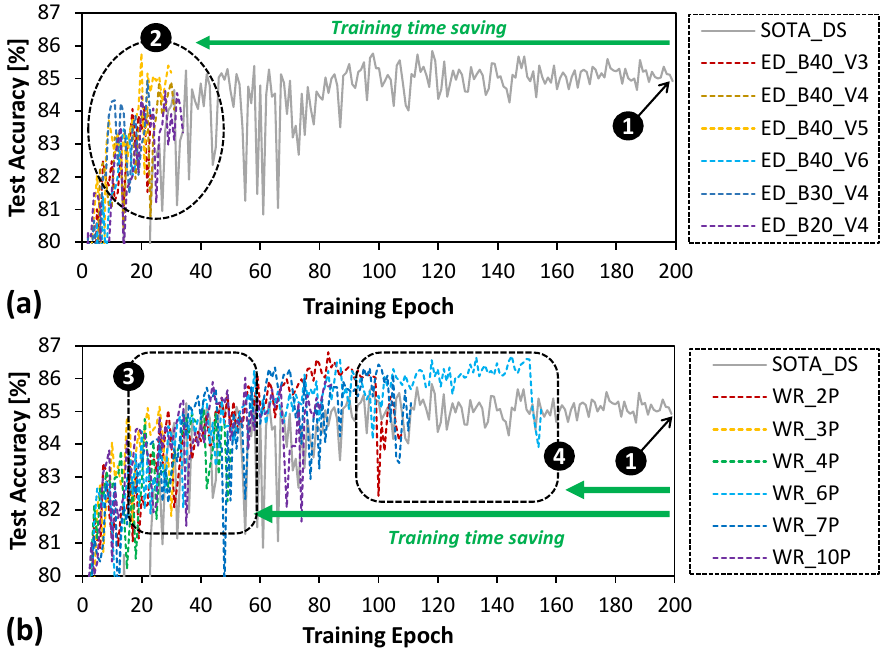}
\vspace{-0.3cm}
\caption{Comparison of accuracy between the SOTA\_DS and our proposed LR policies: (a) exponential decay, and (b) warm restarts.} 
\label{Fig_Results_Accuracy}
\vspace{-0.4cm}
\end{figure}

The experimental results for observing accuracy comparisons between the state-of-the-art (SOTA\_DS) with our proposed exponential decay and warm restarts policies are provided in Fig.~\ref{Fig_Results_Accuracy}(a) and (b), respectively.

\textbf{SOTA\_DS vs. Our Exponential Decay:} 
SOTA\_DS can achieve 84.9\% accuracy after 200 training epochs; see~\circledB{1}. 
If we consider the defined $acc_{th}$, then SOTA\_DS reaches stability after 77 training epochs with 84.7\% accuracy.
Meanwhile, our proposed exponential decay policy can achieve stability faster as compared to SOTA\_DS across different investigated settings, i.e., up to 10.5x compared to SOTA\_DS with full training phase and up to 4x compared to SOTA\_DS with first stable accuracy; see~\circledB{2} and Table~\ref{Tab_DSvsED}.
Besides faster stability, all investigated settings for our exponential decay also lead to comparable accuracy (and better accuracy in some cases) than SOTA\_DS. 
This highlights that our exponential decay policy is effective for learning event-based NCARS dataset.
The reason is that, this policy forces the network to start learning with high LR value (high confidence), then gradually reducing LR for fine-tuning, while considering a selected range of LR values.
Hence, a high LR value in the beginning of the training phase can quickly lead to a global minima as there are only 2 classes to learn from the NCARS dataset. 
Consequently, LR policies that do not start with high LR values usually have significant fluctuation in the early training phase as they are more easily trapped in the local minima; as shown in Fig.~\ref{Fig_AccuracyLRPolicies}(c)-(e). 

\begin{table}[h]
\vspace{-0.2cm}
\centering
\scriptsize
\caption{Comparison between SOTA\_DS vs. our Experimental Decay.} 
\vspace{-0.1cm}
\label{Tab_DSvsED}
\begin{tabular}{l|c|c|c|c|}
\cline{2-5}
 & \textbf{\begin{tabular}[c]{@{}c@{}}Reaching \\ first stable \\ accuracy\\ {[}epochs{]}\end{tabular}} & \textbf{\begin{tabular}[c]{@{}c@{}}Accuracy\\ at first \\ stable \\ accuracy\end{tabular}}           & \textbf{\begin{tabular}[c]{@{}c@{}}Speedup vs. \\ SOTA\_DS\\ full training\end{tabular}} & \textbf{\begin{tabular}[c]{@{}c@{}}Speedup vs. \\ SOTA\_DS\\ first stable \\ accuracy\end{tabular}} \\ \hline
\multicolumn{1}{|c|}{\textbf{\begin{tabular}[c]{@{}c@{}}ED\_B40\_V03\\ ED\_B40\_V04\\ ED\_B40\_V05\\ ED\_B40\_V06\\ ED\_B30\_V04\\ ED\_B20\_V04\end{tabular}}} & \begin{tabular}[c]{@{}c@{}}24\\ 34\\ 31\\ 19\\ 26\\ 35\end{tabular} & \begin{tabular}[c]{@{}c@{}}83.8\%\\ 84.3\%\\ 85.2\%\\ 82.4\%\\ 83.9\%\\ 83.4\%\end{tabular} & \begin{tabular}[c]{@{}c@{}}8.3x\\ 5.9x\\ 6.4x\\ 10.5x\\ 7.7x\\ 5.7x\end{tabular}         & \begin{tabular}[c]{@{}c@{}}3.2x\\ 2.3x\\ 2.5x\\ 4x\\ 3x\\ 2.2x\end{tabular}                \\ \hline
\end{tabular}
\vspace{-0.2cm}
\end{table}

\textbf{SOTA\_DS vs. Our Warm Restarts:}
Our warm restarts policy can achieve stability faster with comparable/higher accuracy than SOTA\_DS with full training phase across different investigated settings, i.e., up to 6.2x; see Table~\ref{Tab_DSvsWR}.
Meanwhile, when compared to SOTA\_DS with first stable accuracy, our warm restarts policy can achieve stability faster with comparable accuracy in some cases (i.e., up to 2.4x faster training; see~\circledB{3}), as well as slower yet higher accuracy in other cases (i.e., up to 85.8\% accuracy; see~\circledB{4}).
The reason is that, warms restarts policy has periodic LR changes, hence having more variability when learning the input features. 
For instance, a high LR value in the beginning of the training phase can quickly lead to a global minima as there are only 2 classes to learn from the NCARS dataset.
Then, decreasing LR helps in the fine-tuning process.
However, in some cases, resetting the LR to a high value and decreasing it fast to a low value may make the optimization sub-optimal and trapped in the local minima, thereby making the network learn slower than SOTA\_DS.

\begin{table}[h]
\vspace{-0.2cm}
\centering
\scriptsize
\caption{Comparison between SOTA\_DS vs. our Warm Restarts.} 
\vspace{-0.2cm}
\label{Tab_DSvsWR}
\begin{tabular}{l|c|c|c|c|}
\cline{2-5}
 & \textbf{\begin{tabular}[c]{@{}c@{}}Reaching \\ first stable \\ accuracy\\ {[}epochs{]}\end{tabular}} & \textbf{\begin{tabular}[c]{@{}c@{}}Accuracy\\ at first \\ stable \\ accuracy\end{tabular}}           & \textbf{\begin{tabular}[c]{@{}c@{}}Speedup vs. \\ SOTA\_DS\\ full training\end{tabular}} & \textbf{\begin{tabular}[c]{@{}c@{}}Speedup vs. \\ SOTA\_DS\\ first stable \\ accuracy\end{tabular}} \\ \hline
\multicolumn{1}{|c|}{\textbf{\begin{tabular}[c]{@{}c@{}}WR\_2P\\ WR\_3P\\ WR\_4P\\ WR\_6P\\ WR\_7P\\ WR\_10P\end{tabular}}} & \begin{tabular}[c]{@{}c@{}}109\\ 32\\ 51\\ 156\\ 112\\ 85\end{tabular}                & \begin{tabular}[c]{@{}c@{}}84.4\%\\ 83.9\%\\ 83.9\%\\ 85.1\%\\ 85.3\%\\ 85.8\%\end{tabular} & \begin{tabular}[c]{@{}c@{}}1.8x\\ 6.2x\\ 3.9x\\ 1.3x\\ 1.8x\\ 2.4x\end{tabular}          & \begin{tabular}[c]{@{}c@{}}0.7x\\ 2.4x\\ 1.5x\\ 0.5x\\ 0.7x\\ 0.9x\end{tabular}            \\ \hline
\end{tabular}
\vspace{-0.2cm}
\end{table}

\vspace{-0.1cm}
\subsection{Reduction of the Carbon Emission}
\label{Sec_Results_Carbon}

\begin{figure}[t]
\centering
\includegraphics[width=\linewidth]{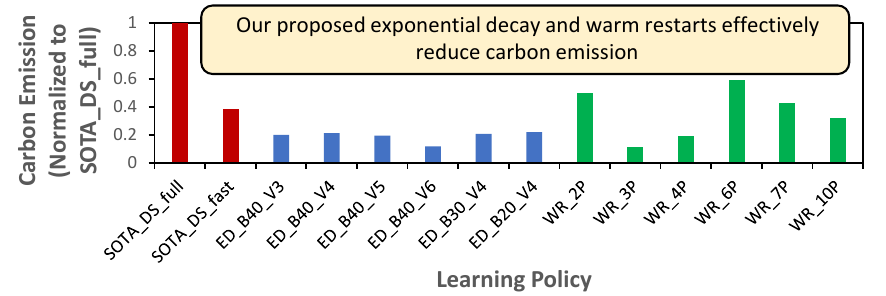}
\vspace{-0.8cm}
\caption{Comparison of carbon emission across different LR policies. Here, SOTA\_DS\_full referes to the SOTA\_DS with full training phase, while SOTA\_DS\_fast refers to the SOTA\_DS with first stable accuracy.} 
\label{Fig_Results_Carbon}
\vspace{-0.5cm}
\end{figure}

Fig.~\ref{Fig_Results_Carbon} shows the results of carbon emission comparison across different LR policies,
The results show that, our proposed LR policies (i.e., exponential decay and warm restarts) effectively reduce the carbon emission as compared to the SOTA\_DS, mainly due to the SNN training time reduction.
When compared to the SOTA\_DS\_full, our proposed LR policies can reduce up to 88.39\% carbon emission.
Meanwhile, when compared to the SOTA\_DS\_fast, our proposed LR policies can reduce up to 69.84\% carbon emission.

\section{Conclusion}
\label{Sec_Conclusion}

We propose a novel FastSpiker methodology to enable fast SNN training on event-based data through learning rate enhancements.
It investigates the impact of different LR policies, then explores different settings for the selected LR policies to find the appropriate policies through a statistical-based decision. 
Experimental results show that our FastSpiker can achieve up to 10.5x faster training time and up to 88.39\% lower carbon emission on the event-based automotive data, hence initiating  the way for green and sustainable computing in realizing embodied neuromorphic intelligence for autonomous embedded systems.


% ...

\section*{ACKNOWLEDGMENT}
\label{Sec_Ack}

This work was partially supported by the NYUAD Center for Interacting Urban Networks (CITIES), funded by Tamkeen under the NYUAD Research Institute Award CG001, and the NYUAD Center for Artificial Intelligence and Robotics (CAIR), funded by Tamkeen under the NYUAD Research Institute Award CG010.
\end{spacing}
\vspace{-0.1cm}

\begin{spacing}{0.99}
\bibliographystyle{IEEEtran}
\bibliography{bibliography}
\end{spacing}

\end{document}